%% file: main.tex
\documentclass{article}
\usepackage{algorithm}
\usepackage{algpseudocode}
\usepackage{amssymb}
\usepackage{amsmath}
\usepackage{graphicx}
\usepackage{multirow}
\usepackage{makecell}
\usepackage{wrapfig}
\usepackage{enumitem}
\usepackage{booktabs}
\setlist[itemize]{nosep,leftmargin=1em,labelsep=0.5em}

\setlist[itemize]{nosep,leftmargin=1em,labelsep=0.5em}

\usepackage[final]{corl_2025} 

\newcommand{\ouralg}{\emph{NeuralSVCD}}
\newcommand{\ourrepsim}{\mathcal{Z}}

\title{NeuralSVCD for Efficient Swept Volume Collision Detection}

\author{%
  Dongwon Son\thanks{Equal contribution.}
  \qquad
  Hojin Jung\footnotemark[\value{footnote}]
  \qquad
  Beomjoon Kim
  \\
  Korea Advanced Institute of Science and Technology (KAIST)\\
  Seoul, South Korea\\
  \texttt{\{dongwon.son, hojin.jung, beomjoon.kim\}@kaist.ac.kr}
}

\begin{document}
\maketitle

\vspace{-7mm}
\begin{abstract}
    Robot manipulation in unstructured environments requires efficient and reliable Swept Volume Collision Detection (SVCD) for safe motion planning. Traditional discrete methods potentially miss collisions between these points, whereas SVCD continuously checks for collisions along the entire trajectory. Existing SVCD methods typically face a trade-off between efficiency and accuracy, limiting practical use.
    In this paper, we introduce \ouralg, a novel neural encoder-decoder architecture tailored to overcome this trade-off. Our approach leverages \emph{shape locality} and \emph{temporal locality} through distributed geometric representations and temporal optimization. This enhances computational efficiency without sacrificing accuracy.
    Comprehensive experiments show that \ouralg\ consistently outperforms existing state-of-the-art SVCD methods in terms of both collision detection accuracy and computational efficiency, demonstrating its robust applicability across diverse robotic manipulation scenarios. Code and videos are available at \url{https://neuralsvcd.github.io/}.
\end{abstract}



\input{intro_v6}

\input{related_works}

\input{method}
\input{experiment}
\input{conclusion}

\clearpage

\input{limitations}


\acknowledgments{
This work was supported by Institute of Information \& communications Technology Planning \& Evaluation (IITP) grant and National Research Foundation of Korea (NRF) funded by the Korea government(MSIT) (No.2019-0-00075, Artificial Intelligence Graduate School Program(KAIST)), (No.2022-0- 00311, Development of Goal-Oriented Reinforcement Learning Techniques for Contact-Rich Robotic Manipulation of Everyday Objects), (No. 2022-0-00612, Geometric and Physical Commonsense Reasoning based Behavior Intelligence for Embodied AI), (No. RS-2024-00359085, Foundation model for learning-based humanoid robot that can understand and achieve language commands in unstructured human environments), (No. RS-2024-00509279, Global AI Frontier Lab).
}


\bibliography{ref}  

\appendix
\input{Appendix}

\end{document}

%% file: intro_v6.tex
\begin{figure}[h]
    \vspace{-2mm}
    \centering
    \includegraphics[width=0.8\linewidth]{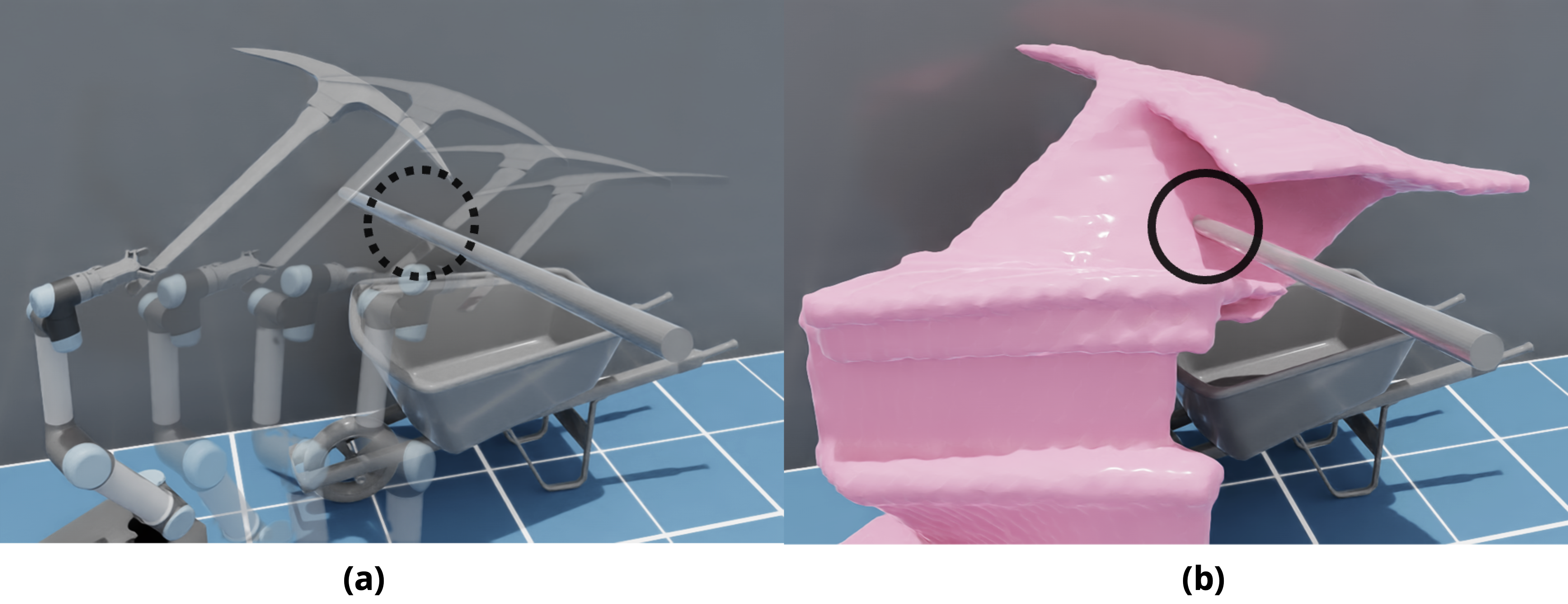}
    \caption{(a) Illustration of tunneling errors in discrete collision detection. Sampling only a finite set of waypoints along the robot's trajectory can miss collisions occurring between waypoints (highlighted by the dashed circle). (b) Swept Volume Collision Detection (SVCD). SVCD evaluates the collision between the swept volume of the object along the given trajectory (pink) and obstacles.}
    \label{fig:ccd_importance}
    \vspace{-3mm}
\end{figure}

\section{Introduction}
\begin{figure}
    \centering
    \includegraphics[width=0.9\linewidth]{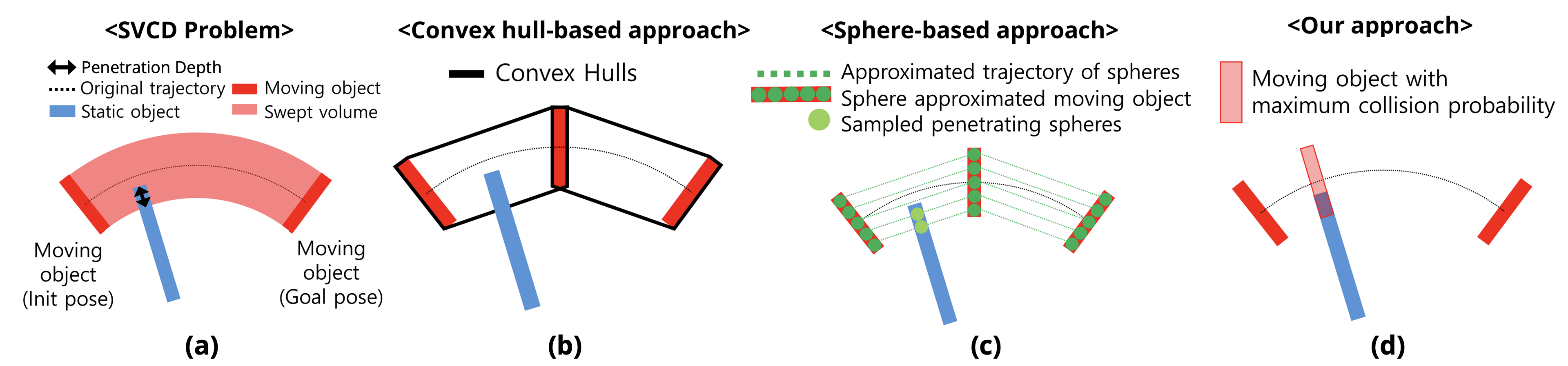}
    \caption{
    (a) SVCD problem definition. (b-d) Comparison of geometric representations used in different SVCD algorithms. 
    }
    \label{fig:baseline_comparison}
    \vspace{-3mm}
\end{figure}

\vspace{-3mm}
Robotic motion planning in unstructured environments, such as inserting dishes into a rack, assembling parts, or navigating in a tight space, requires accurate collision checking.
Traditional discrete collision checkers sample a finite set of waypoints along a candidate path, but this can overlook collisions that occur between samples, known as ``tunneling'' errors. In contrast, SVCD prevents tunneling problems by using the swept volume of a path of the object to check collision, making it more practical, especially in tight‐tolerance tasks~\cite{ericson2004realtime}. See Figure \ref{fig:ccd_importance}. Formally, given an object (or robot) trajectory \(\tau:[0,1]\to SE(3)\) and geometric models of the objects, SVCD algorithms output the maximum penetration depth. See Figure \ref{fig:baseline_comparison}, (a) for an illustration.

SVCD algorithms typically have two design components: the geometric representation of entities and a collision detector tailored for the representation. For instance, previous works~\cite{xavier1997fast, schulman2013finding} discretize trajectories, construct convex hulls using vertices of a pair of waypoints in the trajectory, then apply the Gilbert–Johnson–Keerthi (GJK) algorithm~\cite{gilbert2002GJK} designed for convex shapes to each convex hull (Figure \ref{fig:baseline_comparison}, (b)). Ideally, we would parallelize the collision checks on convex hulls using a GPU, but because GJK involves branching in computation, this is difficult and it must iterate through each convex hull to check collisions~\cite{son2023local}, which makes it inefficient. In contrast, collision detection between spheres can be computed in parallel on a GPU, as it only involves evaluating distances between sphere centers. Recent approaches~\cite{sundaralingam2023curobo, ramsey2024collision, sui2024hardware} exploit this property to approximate complex shapes with multiple spheres, discretize trajectories at a high resolution, and use parallelized sphere-based collision checks at all waypoints (see Figure \ref{fig:baseline_comparison} (c)). However, this approximation sacrifices accuracy, leading to inaccurate collision detection.

Instead, we propose a neural network (NN)-based SVCD algorithm using an encoder-decoder architecture. Specifically, the encoder processes pairs of object shapes, poses, and trajectory data to produce neural representations of the shapes, while the decoder predicts collision outcomes directly from these representations. This offers two primary advantages. First, by training an encoder-decoder network for collision prediction, the encoder learns a latent representation that preserves only the relevant features, rather than a complete mesh. Second, since the decoder is a NN, it can leverage GPU parallelization. However, there are two challenges involved in this approach. First, we have a limited amount of 3D asset data~\cite{downs2022google, deitke2023objaverse}, but we need to make the model generalize across a variety of novel shapes. Second, even for the same shapes, you can create a large variety of object trajectories, where even seemingly similar trajectories could have different collision outcomes, especially in tight spaces.

\begin{wrapfigure}{r}{0.5\textwidth}
\vspace{-4mm}
\begin{center}
    \includegraphics[width=0.5\textwidth]{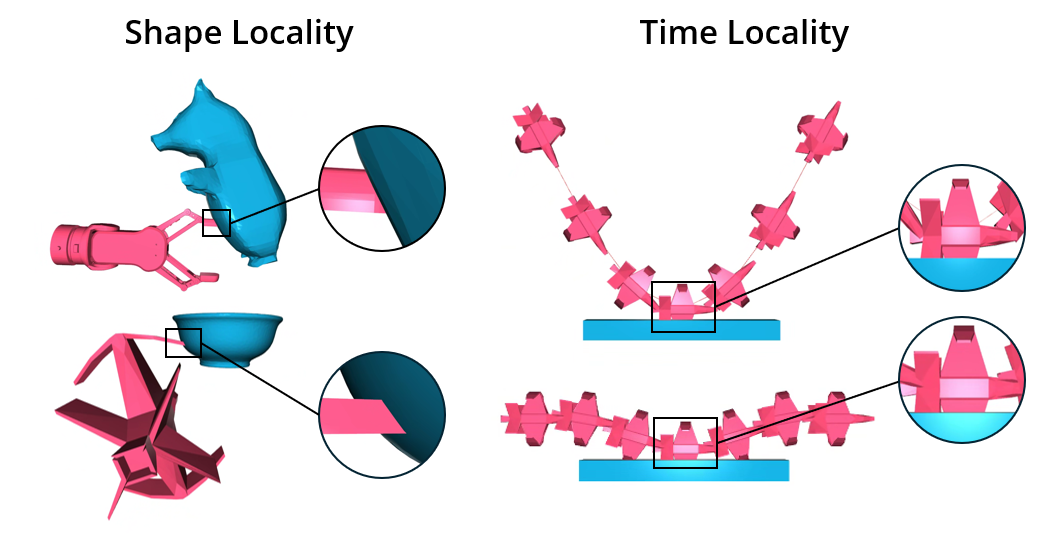}
    \end{center}
    \caption{
    (Left - shape locality) Two different pairs of objects have completely different global shapes, but when we focus on the circled contact regions, they have very similar shapes. (Right - temporal locality) Two different object trajectories share the same collision moment, marked by rectangles.
    }
    \label{fig:locality}
\end{wrapfigure}
To overcome these limitations, we leverage two essential insights: \emph{shape locality} and \emph{temporal locality}. Shape locality addresses the scarcity of 3D asset data by recognizing that local geometric features remain relatively consistent, even when global shapes differ significantly, as shown in Figure \ref{fig:locality} (Left). So, if we restrict the neural decoder's input to local object features that are relevant for collision rather than global object geometry, the model would generalize collision predictions across unseen shapes. Temporal locality is the observation that collisions are generally influenced by short trajectory segments rather than an entire path, as depicted in Figure \ref{fig:locality} (right). Consequently, if we limit the decoder's input to these relevant trajectory segments instead of complete trajectories, our approach would be more robust across diverse trajectories.

Based on these insights, we propose \ouralg, which uses a novel geometric representation designed to leverage shape locality and temporal locality. Specifically, we represent a shape as a set of uniformly sampled points on the object and encode the local shape around each point into a latent representation using a neural network. The entire shape, then, is represented with a collection of these local latent representations. A decoder, instead of taking the entire object shapes as inputs, only takes a pair of these local representations to check if there is a collision between these two local shapes. However, naively checking a collision this way would require evaluating $N^2$ pairs of local representations continuously over the trajectory, where $N$ is the number of local representations.
To avoid excessive collision checks, we propose instead to first perform a broad-phase collision detection by approximating local shapes with spheres.
Then, we transform the spheres along the trajectory to find the time and pair of spheres that intersect.
This process yields the specific pair of local latent vectors and the corresponding collision time along the trajectory.
Then, we run the collision decoder on pairs whose corresponding spheres are in collision, supplying the corresponding locally linearized trajectory segment.
This allows us to limit the decoder's input to these relevant trajectory segments and local shapes rather than complete trajectories and global shapes.



We benchmark our approach against multiple state-of-the-art SVCD baselines, including mesh-based and sphere-based methods, demonstrating significant improvements in both swept volume collision detection accuracy and runtime.
When integrated into the cuRobo trajectory optimizer and evaluated on three challenging robotics tasks, \ouralg\ achieves higher success rates and reduced computation times compared to state-of-the-art approaches.



%% file: related_works.tex
\vspace{-3mm}
\section{Related works}
\vspace{-3mm}

\subsection{Explicit shape representations for swept volume collision detection}
\vspace{-2mm}
Continuous collision detection (CCD) and SVCD both answer whether a moving object intersects a static obstacle along a given trajectory. CCD additionally returns the time of first contact (TOC),  which is the first time at which the two meshes collide \cite{canny1986collision,kim2003collision,redon2000algebraic,tang2009c2a}. SVCD instead outputs the maximum penetration depth encountered over the entire motion. SVCD involves two critical challenges: (i) reconstructing the swept volume of the moving object and (ii) accurately measuring its penetration depth against the static mesh.

There are lines of work that reconstruct swept volume as triangle mesh \cite{peternell2005swept, rossignac2007boundary, abrams2000computing, kim2003fast,zhang2009reliable}.
Even though they can generate accurate swept volumes, computing penetration depth against a non-convex mesh of the static object is slow due to exhaustive triangle–edge pair tests. To accelerate queries, an alternative approach forgoes exact swept-volume reconstruction and instead approximates the swept volume with a collection of convex hulls~\cite{schulman2013finding,xavier1997fast}, which enables use of GJK~\cite{gilbert2002GJK}.
We obtain the swept volume by convex‐decomposing both object meshes, discretizing the trajectory into segments, and for each segment and convex part, building the convex hull of its vertices at that segment’s start and end poses (see Fig.~\ref{fig:baseline_comparison}, second column).
After reconstructing the swept volume, the penetration depth is calculated by applying the GJK algorithm for each pair of convex cells from different objects.
This hull-based approach is exact only under pure translation, and GJK’s iterative, branch-heavy algorithm hinders parallel performance, making it inefficient for GPU-accelerated pipelines.

An alternative for efficient penetration‐depth estimation decomposes the moving object into a set of spheres~\cite{sundaralingam2023curobo}. In CuRobo, these spheres query signed distances against the static mesh in parallel on the GPU. Penetration is measured by discretizing the trajectory into piecewise‐linear segments, selecting all spheres that intersect the environment along each segment, and summing their penetration depths (Fig.~\ref{fig:baseline_comparison}, third column).
Although this approach scales linearly with the number of spheres, its geometric fidelity suffers for thin or highly concave shapes, as accurately covering narrow features or deep cavities requires a prohibitively large number of spheres, as demonstrated in our motion planning experiments.

\vspace{-1mm}
\subsection{Implicit shape representations for swept volume collision detection}
\vspace{-2mm}
To support general trajectories, recent works use the implicit function, which is a function that outputs the degree of distance to the surface from a given point, taking negative values within the object and positive values outside. The swept volume is represented by an implicit function that, for any point in space, takes the smallest value of the object’s original implicit function observed over the entire motion, evaluated with respect to time along the trajectory \cite{sellan2021swept}. This process is referred to as time optimization because it determines the timestep at which maximum penetration occurs between a point and a moving object along its trajectory.
Penetration depth between the swept volume and the static object is computed by evaluating the swept volume’s implicit function on the static object’s surface and taking the pointwise minimum.
This method supports arbitrary trajectories and any implicit representation. However, accurate penetration‐depth estimation via implicit functions requires dense surface sampling and a separate time‐optimization solution for each sample, which becomes prohibitively expensive.


Instead, we can use an object collision network, which predicts contact probability based on the shape and pose of two objects \cite{son2023local, son2020sim, danielczuk2021object}, rather than an implicit function, to accelerate the time optimization process.
Time optimization, which originally finds the moment of maximal penetration between a moving object and a single surface point, can be reframed as locating the time of maximum penetration between the moving mesh and an entire static object (Fig.~\ref{fig:baseline_comparison}, fourth column). To our knowledge, ours is the first SVCD approach to fuse an object-collision network into this time-optimization loop for defining the swept volume.


%% file: method.tex
\vspace{-2mm}
\section{Neural swept volume collision detection}
\label{sec:method}
\vspace{-2mm}

\textit{Swept volume collision detection} (SVCD) involves identifying potential collisions between a static environment and the swept volumes generated by objects moving along specified trajectories. Formally, the \textit{swept volume} (SV) of an object is defined as the spatial region occupied by its geometry as it continuously moves along a trajectory \(\tau:[0,1]\rightarrow SE(3)\). Given meshes for both a static scene (stationary environment) and an object at a canonical pose, SVCD determines if there is a collision between the object moving through $\tau$ and the scene.
Throughout this paper, the subscript \(mesh_{static}\) denotes a representation of the mesh of a static object that remains fixed during the collision detection process, and the subscript \(mesh_{mov}\) specifically indicates representation of a moving object at its canonical pose, prior to their transformation along trajectory \(\tau\).


\begin{figure}
\vspace{-8mm}
    \centering
    \includegraphics[width=1.0\linewidth]{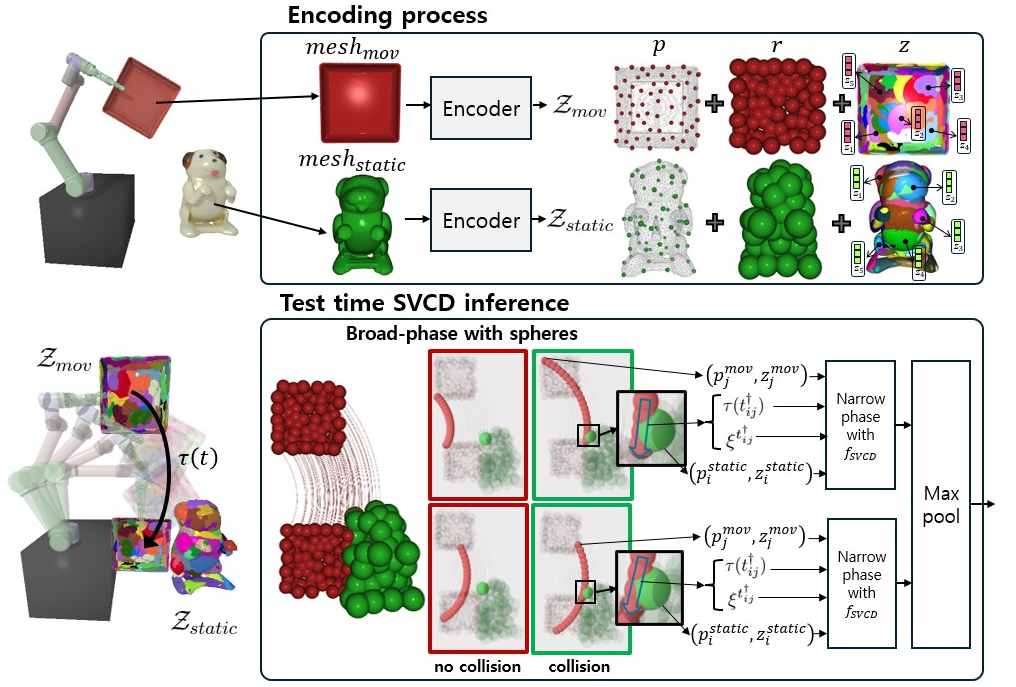}
    \caption{
    Overview of SVCD pipeline.
    During encoding, (top), the canonical meshes (\(\text{mesh}_{\text{mov}}\) and (\(\text{mesh}_{\text{static}}\))  transform into distributed latent representations (\(\ourrepsim_{\text{mov}}\) and \(\ourrepsim_{\text{static}}\), respectively) using a neural encoder.
    Each representation \(\ourrepsim\) consists of \(N\) representative points (\(p^{\text{static}}_i\) and \(p^{\text{mov}}_j\)), associated bounding spheres with radii ($r_i^{static}$ and $r_j^{mov}$), and local latent vectors (\(z^{\text{static}}_i\) and \(z^{\text{mov}}_j\)).
    During inference (bottom), given latent representations \(\ourrepsim_{\text{mov}}\) and \(\ourrepsim_{\text{static}}\) along with a trajectory \(\tau(t):[0,1]\rightarrow SE(3)\) (bottom left), we initially evaluate all possible pairs \(\{(i,j) \mid i,j \in [1,N]\}\).
    In the broad phase (middle), we use bounding spheres to quickly identify intersecting pairs \((i,j)\) and its time \(t^{\dagger}_{ij}\) of maximum sphere overlap. The local trajectory is defined as the linearized motion at \(\xi^{t^{\dagger}_{ij}} \in \mathbb{R}^6\) by computing its first-order linear approximation (see black box).
    In the narrow-phase (right), for all identified collision candidate pairs and time \((i,j,t^{\dagger}_{ij})\), the neural SVCD decoder \(f_{SVCD}\) refines collision predictions based on inputs \((p_i^{\text{static}},z_i^{\text{static}})\), \(\xi^{t^{\dagger}}\), \(\tau(t^{\dagger})\), and \((p_j^{\text{mov}},z_j^{\text{mov}})\). The final collision outcomes for rigid bodies are aggregated using max-pooling across all identified pairs.
    }
    \label{fig:inference_flow}
\end{figure}

\vspace{-2mm}
\subsection{Distributed latent scene representation from neural encoder}
\label{sec:distributed_latent_scene_rep}
\vspace{-2mm}
Given meshes \(\text{mesh}_{\text{mov}}\) and \(\text{mesh}_{\text{static}}\), our first step is to encode these meshes into distributed latent object representations \(\ourrepsim\). To leverage shape locality effectively, we adopt a distributed latent representation strategy. Specifically, we first sample surface points from each mesh and select \(N\) representative points \(\{p_i\}_{i=1}^N\) using furthest point sampling (FPS), ensuring uniform coverage across the surface. Each sampled surface point is then assigned to its nearest representative point \(p_i\). For each representative point \(p_i\), we input its assigned local points into a neural network to generate a localized latent shape representation \(z_i \in \mathbb{R}^K\). This process yields distributed pairs \(\{(p_i,z_i)\}_{i=1}^N\).

For efficient broad-phase collision detection, we define bounding spheres around each representative point \(p_i\) with radius \(r_i \in \mathbb{R}\). 
We define the radius \(r_i\) for each point based on the distance to its nearest neighboring representative point: \(
r_i = \alpha \min_{j \in [1,N], j \ne i} \text{dist}(p_i, p_j)
\), where $\alpha > 1$, ensuring each sphere encloses the local geometry near its representative point. Consequently, our final distributed latent object representation is \( \ourrepsim = \{(p_i, z_i, r_i)\}_{i=1}^N \), illustrated in the upper part of Figure~\ref{fig:inference_flow}.

\vspace{-2mm}
\subsection{Two-Step SVCD via Broad-phase Filtering and Neural Refinement}
\vspace{-2mm}
Given distributed representations of the moving object \( Z_{\text{mov}} = \{(p_i^{\text{mov}}, z_i^{\text{mov}}, r_i^{\text{mov}})\}_{i=1}^{N} \) at the initial trajectory position \( \tau(0) \), static scene \( Z_{\text{static}} = \{(p_j^{\text{static}}, z_j^{\text{static}}, r_j^{\text{static}})\}_{j=1}^{N} \), and a trajectory \( \tau: [0,1] \rightarrow SE(3) \), we propose a two-step SVCD approach to improve computational efficiency. Our method comprises a \emph{broad-phase} that quickly identifies potential collision candidates, followed by a \emph{narrow-phase} employing a neural SVCD decoder for precise collision evaluation.

\textbf{Broad-phase with Sphere Approximation:}
Performing collision detection between the static object and the swept volume of one moving object typically involves checking all possible pairs of local representations, leading to \(N^2\) checks continuously over the trajectory, which is computationally intensive. To reduce complexity, we first use a broad-phase collision detection to identify potential collision pairs and their corresponding times efficiently.
Specifically, we use bounding spheres with radii \( r_i^{\text{static}} \) for the static object and \( r_j^{\text{mov}} \) for the moving object.
Throughout this section, the subscript \( \cdot_i \) denotes elements of the static representation \(\ourrepsim_{\text{static}}\), and the subscript \( \cdot_j \) denotes elements of the moving representation \(\ourrepsim_{\text{mov}}\).
We evaluate potential collisions by checking overlaps between each static sphere and the corresponding swept volume formed by each moving sphere as it follows the trajectory \(\tau(t)\).
For each pair \((i,j)\) within the complete set \(\mathcal{P}_{\text{all}} = \{(i,j) \mid i,j \in [1,N]\}\), we perform the following numerical optimization (e.g., via Newton's method):
\begin{equation}
\label{eq:broad_phase_dist}
\min_{t \in [0,1]} \text{dist}(p_i^{\text{static}}, \tau(t) \cdot p_j^{\text{mov}}) - r_i^{\text{static}} - r_j^{\text{mov}},
\end{equation}
where \(\text{dist}(\cdot, \cdot)\) denotes the Euclidean distance between points, and \(\tau(t) \cdot p_j^{\text{mov}}\) represents the transformed position of \(p_j^{\text{mov}}\) under the trajectory \(\tau(t) \in SE(3)\).
We denote by \(t^\dagger_{ij}\) the time along the trajectory at which this quantity is minimized—that is, the moment when the two bounding spheres are in closest proximity.
Negative results from Eq.~\eqref{eq:broad_phase_dist} indicate possible collisions, forming the reduced candidate set of \((i,j,t^\dagger_{ij})\), which requires further detailed evaluation in the \emph{narrow-phase}. Additionally, solving this optimization provides the argmin solution—the pseudo-critical collision time \( t^{\dagger} \)—which serves as an approximation to the true critical collision time \( t^* \), derived from simplified spherical approximations rather than exact geometrical representations.

\textbf{Narrow-phase with Neural Decoder:}
For each candidate pair \((i,j,t^{\dagger}_{ij})\) identified in the broad-phase, we refine collision predictions using a neural SVCD decoder \(f_{SVCD}\). The decoder takes as input the static representation \((p_i^{\text{static}}, z_i^{\text{static}})\) and the swept volume representation generated by \((p_j^{\text{mov}}, z_j^{\text{mov}})\) along the trajectory \(\tau\). To exploit temporal locality, we avoid using the entire trajectory, focusing instead on a local trajectory segment around \(t^{\dagger}_{ij}\), we approximate the local trajectory using a linear, first-order Taylor expansion parameterized by \(\xi^{t^{\dagger}_{ij}} \in \mathbb{R}^6\), representing linear and angular velocities at \(t^{\dagger}_{ij}\), along with the pose \(\tau(t^{\dagger}_{ij})\). We define the local swept volume segment using \((p_j^{\text{mov}}, z_j^{\text{mov}})\) transformed by this local trajectory approximation. The neural decoder \(f_{SVCD}\) then processes inputs \((p_i^{\text{static}}, z_i^{\text{static}})\), \((p_j^{\text{mov}}, z_j^{\text{mov}})\), \(\xi^{t^{\dagger}_{ij}}\), and \(\tau(t^{\dagger}_{ij})\) to predict the collision probability.

We assume that the pseudo-critical collision time \(t^\dagger_{ij}\) sufficiently approximates the true critical collision time \(t^*\), which is time where decoded collision probability is maximized, ensuring that the local swept volume defined by \(\xi^{t^\dagger_{ij}}\) and \(\tau(t^\dagger_{ij})\) adequately captures the critical collision region for accurate predictions.
Specifically, the neural decoder \(f_{SVCD}\) receives the inputs \((p_i^{\text{static}}, z_i^{\text{static}})\), \((p_j^{\text{mov}}, z_j^{\text{mov}})\), \(\xi^{t^\dagger_{ij}}\), and \(\tau(t^{\dagger}_{ij})\). We train \(f_{SVCD}\) to directly predict collision probabilities from these inputs. Practically, \(f_{SVCD}\) consists of multiple multilayer perceptrons (MLPs) that process concatenated inputs and output collision probabilities within \([0,1]\). Final collision predictions for rigid bodies are obtained by applying max pooling across all candidate pair probabilities identified during the broad-phase. This inference procedure is illustrated in the lower portion of Figure \ref{fig:inference_flow}.

\vspace{-2mm}
\subsection{Training Encoder and Decoder}
\label{sec:training}
\vspace{-2mm}
In this section, we describe our procedure for training both the shape encoder and the neural SVCD decoder.
Our training dataset consists of 2556 carefully curated meshes sourced from established 3D asset collections, including GoogleScannedObjects \cite{downs2022google}, ObjaverseXL \cite{deitke2023objaverse}, and NOCS \cite{Wang_2019_CVPR}.
We create training datapoints by randomly pairing a static mesh \(\text{mesh}_{\text{static}}\) with a canonical moving mesh \(\text{mesh}_{\text{mov}}\). These pairs undergo random transformations, and we sample trajectories \(\tau(\cdot):[0,1]\rightarrow SE(3)\). For training simplicity, we restrict trajectories to linear paths by randomly selecting directions in \(\mathbb{R}^6\) and an initial pose \(\tau(0) \in SE(3)\). Ground-truth collision labels \(y_{SVCD}\) are computed using the FCL library \cite{pan2012fcl}. Thus, each datapoint is formally defined as \(d := (\text{mesh}_{\text{static}}, \text{mesh}_{\text{mov}}, \tau(\cdot), y_{SVCD})\). To ensure balanced training, we apply rejection sampling to maintain an equal proportion of collision and non-collision instances, resulting in our final training dataset \(\mathcal{D}\).
The training objective is defined as \(\sum_{d\in\mathcal{D}} L_{SVCD}(d),\)
where \(L_{SVCD}\) represents the binary cross-entropy loss between the predicted collision outcomes and the ground-truth labels \(y_{SVCD}\). To enable NN to output the degree of penetration instead of the predicted binary label, we add a regularization loss. See Appendix \ref{app:train_detail} for further details.

%% file: experiment.tex
\vspace{-2mm}
\section{Experiments}
\label{sec:experiment}
\vspace{-2mm}
Our experiments are designed to test two main hypotheses: (1) \ouralg\ outperforms the state-of-the-art methods for SVCD in terms of both accuracy and computational efficiency. (2) When integrated into a motion planning framework, our SVCD algorithm leads to higher success rates and better efficiency compared to systems employing the baseline SVCD implementations.

 \begin{figure}
 \vspace{-4mm}
    \centering
    \includegraphics[width=1.0\linewidth]{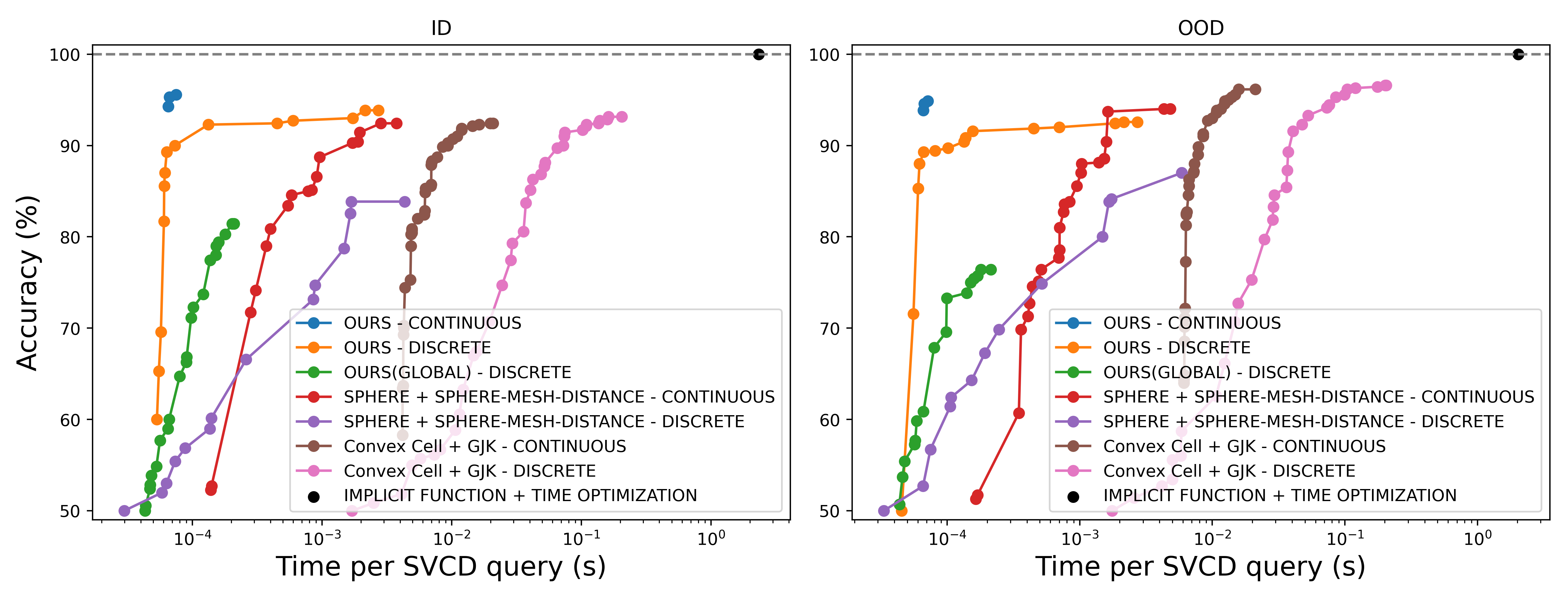}
    \caption{Accuracy efficiency tradeoff graph of our SVCD and other baselines for (Left) in-domain object sets and (Right) out-of-domain object sets.}
    \label{fig:ccd_accuracy}
    \vspace{-2mm}
\end{figure}

\vspace{5mm}
\subsection{Swept Volume Collision Detection Accuracy}
\label{sec:exp_SVCD_acc}
\vspace{-2mm}
\begin{wrapfigure}{r}{0.5\textwidth}
    \vspace{-4mm}
    \centering
    \includegraphics[width=0.5\textwidth]{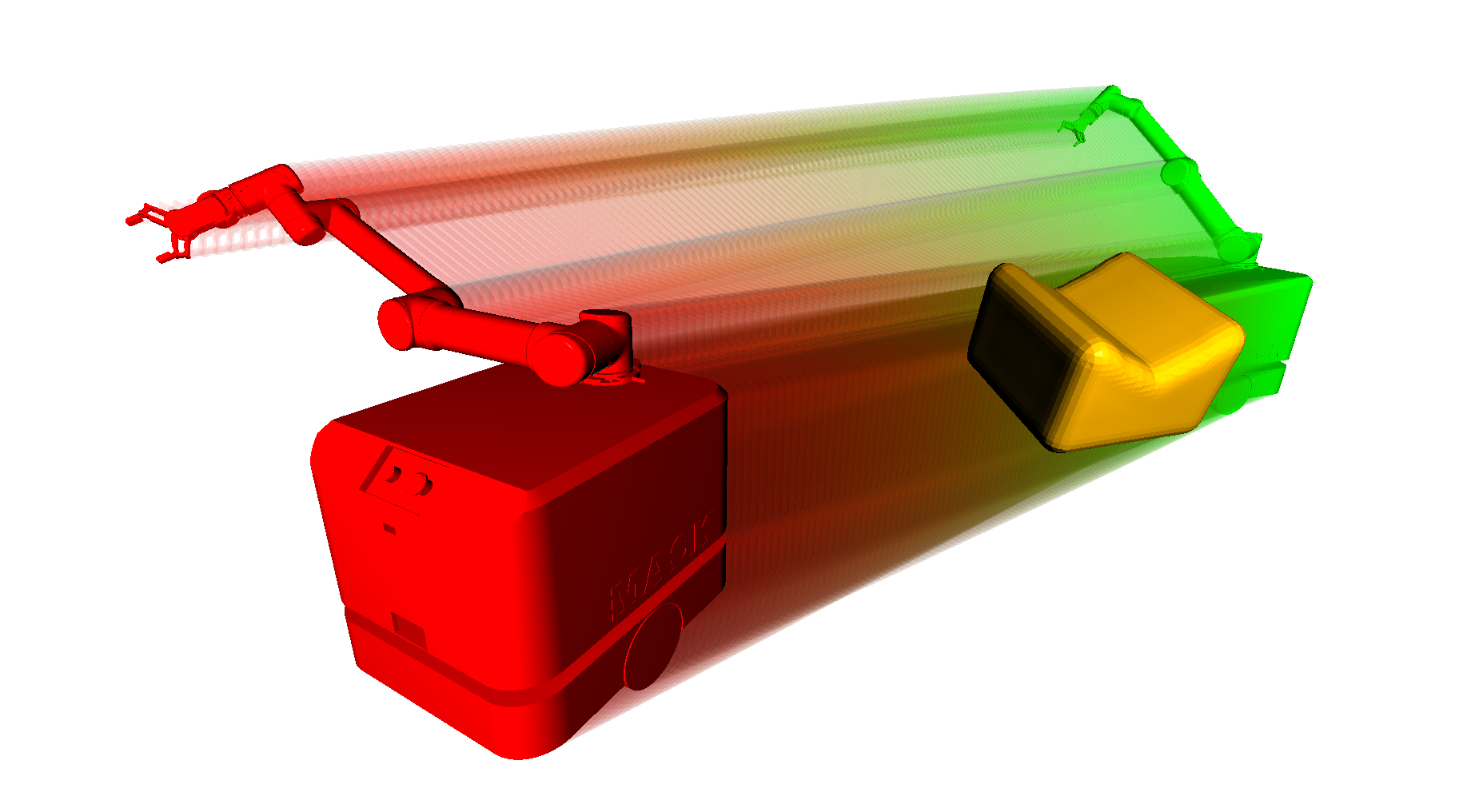}
    \caption{
    This figure illustrates the problem case for checking swept volume collision accuracy. The objective of each problem is to determine where a moving robot, starting from the initial configuration (marked in green) and moving to the goal configuration (marked in red), will collide with a static object (marked in yellow). 
    }
    \label{fig:ccd_problem}
    \vspace{-2mm}
\end{wrapfigure}

To quantify the accuracy of the SVCD, we perform a comparative experiment. We use $9$-DOF UR5 robot with mobile base as moving robot (3-base, 6-arm). For each trial, we randomly assign the initial and goal configurations within a bounded workspace and select a static object from a predefined pool, assigning it a random pose. Visualization of the problem case is shown in Figure~\ref{fig:ccd_problem}. To evaluate generalization, we partition the obstacles into two asset pools: \textbf{In‑Domain (ID)}—objects that were present during training—and \textbf{Out‑of‑Domain (OOD)}—novel shapes unseen at training time. Near-contact situations are critical for collision-free planning in confined spaces. As in \cite{son2020sim}, we adjust the pose of the static object to ensure a small penetration depth. A detailed problem-generation process is described in the Appendix \ref{app:svcd_detail}.


We compare \ouralg\ with seven baselines:
\begin{itemize}
    \item \textbf{Convex Cell/GJK:} Following \cite{schulman2013finding, xavier1997fast}, we convex-decompose moving and static meshes. The \emph{continuous variant} discretizes trajectories into segments and forms convex hulls of each moving object part at segment endpoints. The \emph{discrete variant} unions convex parts at discrete waypoints. Both use GJK collision detection on convex cell pairs.
    \item \textbf{Sphere/Sphere-Mesh Distance:} As in \cite{sundaralingam2023curobo}, the moving object mesh is approximated by spheres (Appendix \ref{app:svcd_detail}). The \emph{continuous variant} samples sphere paths along trajectory segments, while the \emph{discrete variant} samples at waypoints. Collision detection involves sphere–mesh distance queries.
    \item \textbf{Implicit‐function/time optimization:}
    Following \cite{sellan2021swept}, we define the moving object by its signed‐distance implicit function and densely sample surface points on the static mesh. For each point, we solve a time-optimization problem to determine the timestamp at which the trajectory achieves maximum penetration. This method produces accurate ground-truth collision labels, provided that a sufficient number of surface points are sampled.
    \item \textbf{\ouralg\ - Discrete:} 
    This variant uses the same encoder-decoder but replaces time optimization with fixed waypoint sampling: the trajectory is approximated by a set of discrete configurations, and collision checks are evaluated only at those points.
    \item \textbf{\ouralg\ - Global Representation:} This variant replaces multiple local latent representations \( z \) with a single global latent representation.
\end{itemize}

We randomly sample hyperparameters (e.g., interpolation length; see Table \ref{tab:hyperparam_ranges}) and measure collision-detection accuracy and per-query inference time.
Figure \ref{fig:ccd_accuracy} shows that \ouralg\ outperforms all baselines: it attains $90\%$ accuracy in $6.4\times10^{-5}\,\mathrm{s}$ and peaks at $95.6\%$ (ID) and $94.9\%$ (OOD).
In contrast, convex-cell/GJK continuous and discrete variants are $120\times$ and $590\times$ slower (due to GPU-unfriendly branching), while sphere-mesh methods require $24\times$ more time in the continuous case and never reach $90\%$ in the discrete case (owing to the large number of spheres and interpolations).
\ouralg\ using global representation achieves only $81\%$ (ID) and $76\%$ (OOD) of maximum success rate.
\ouralg’s discrete variant degrades maximum success rate by $1.7\%$ (ID) and $2.3\%$ (OOD), and 5\% lower success rate at same inference time of $6.4 \cdot 10^{-5}$s in both ID and OOD settings because grazing contacts require finer discretization.

\begin{table}[t]
\centering
\scriptsize
\renewcommand\theadalign{tl} 
\begin{tabular}{c c l c c c}
\hline
\textbf{Task} & \textbf{Robot} & \textbf{Motion planner} & 
\textbf{Suc Rate} (\%) & 
\textbf{Time} (s) & 
\textbf{Max Pen} (mm) \\
\hline
\multirow{4}{*}{\makecell[c]{dish\\insertion}} 
    & \multirow{4}{*}{\makecell[c]{UR5\\ (6 DoF)}} 
    & cuRobo-sphere-50   & 73.6   & 9.99   & 3.73  \\
\cline{3-6}
 &   & cuRobo-sphere-400  & 95 & 12.35 & 0.5 \\
\cline{3-6}
 &   & cuRobo-\ouralg-discrete   & 8.5 & 26.62 & 52.13 \\
\cline{3-6}
 &   & cuRobo-\ouralg                   & \textbf{99.3}   & \textbf{9.64}   & \textbf{0.01} \\
\hline
\multirow{4}{*}{\makecell[c]{bimanual\\insertion}} 
    & \multirow{4}{*}{\makecell[c]{ARMADA~\cite{kim2025design}\\ (12 DoF)}}
    & cuRobo-sphere-100   &  70.4  & 5.27  & 3.64 \\
\cline{3-6}
 &   & cuRobo-sphere-1200 & 84.9 & 12.45 & 1.03  \\
\cline{3-6}
 &   & cuRobo-\ouralg-discrete & 82.2 & 3.26 & 2.72  \\
\cline{3-6}
 &   & cuRobo-\ouralg                   & \textbf{92.8}   & \textbf{2.92}      & \textbf{0.33}  \\
\hline
\multirow{4}{*}{\makecell[c]{mining\\site\\navigation}} 
    & \multirow{4}{*}{\makecell[c]{mobile UR5 \\ (9 DoF)}} 
    & cuRobo-sphere-50   & 73.8    & 11.67      & 10.82     \\
\cline{3-6}
 &   & cuRobo-sphere-2000     & 83.8    & 60.91      & 5.32     \\
\cline{3-6}
 &   & cuRobo-\ouralg-discrete      & 1.0    & 11.92      & 86.15     \\
\cline{3-6}
 &   & cuRobo-\ouralg                   & \textbf{92.2}   & \textbf{9.84}   & \textbf{3.81}  \\
\hline
\end{tabular}
\caption{
Motion planning performance over 500 random trials per task, reporting average success rate (Suc Rate), planning time (Time), and maximum penetration depth (Max Pen).
}
\vspace{-4mm}
\label{tab:mp-performance-results}
\end{table}

\vspace{-2mm}
\subsection{Motion planning}
\vspace{-2mm}


To demonstrate improvements in motion planning performance, we integrate \ouralg\ into the cuRobo trajectory optimization framework~\cite{sundaralingam2023curobo}, which we refer to as cuRobo-\ouralg.
Given meshes of the static scene, robot links, grasped objects, and specified start and goal robot configurations, the motion planning algorithm generates collision-free trajectories between these configurations.
Initially, we represent the trajectory as a spline with a fixed number of control points.
The initial trajectory is a linear interpolation between the initial and goal states.
We then apply an optimization-based motion planner with an objective function from cuRobo but with our custom collision cost. 

We compare two baselines that use the same planner but different collision detectors. The first, cuRobo-sphere, employs sphere approximations for SVCD; we test it with varying sphere counts per task (denoted cuRobo-sphere-{number of spheres}) to explore the accuracy–efficiency trade-off. The second, cuRobo-\ouralg-discrete, uses our discrete SVCD variant (Section \ref{sec:exp_SVCD_acc}), approximating trajectories by fixed waypoints and performing collision checks only at those configurations.

We evaluate performance on three distinct motion‐planning tasks—dish insertion, bimanual peg assembly, and mining‐site navigation—as illustrated in Figure \ref{fig:mp_task_visualization}.
These tasks, which require navigating tight spaces and interacting with shallow geometries, underscore the necessity of SVCD throughout the entire trajectory.
Each task is defined by initial and goal configurations for the robot, a set of robot links, a set of static obstacles, and hyperparameters for the optimization, such as a limit on the number of optimization iterations. For a detailed description, see \ref{app:mp_domain}.

We evaluate performance by measuring success rate, execution time, and maximum penetration depth over 500 different problem instances for each task.
A task is considered successful if no collisions occur along the entire trajectory and the goal is reached.
The aggregated results are presented in Table \ref{tab:mp-performance-results}.
Across all three tasks, cuRobo-\ouralg\ consistently achieves higher success rates, faster runtimes, and lower maximum penetration depths than both cuRobo‐sphere and cuRobo‐\ouralg‐discrete. Although increasing sphere counts in cuRobo-sphere can match our success rate, it incurs at least $1.3\times$ and up to $6.2\times$ longer computation time.
Adequate on bimanual insertion, the discrete variant falls below $10\%$ in complex or large‐scale scenarios, whereas our continuous approach maintains a robust $92.2\%$ success. These results highlight \ouralg’s versatility and safety across diverse robot platforms and environments (see Appendix \ref{app:mp_detail}).

%% file: conclusion.tex
\section{Conclusion}
\label{sec:conclusion}

We introduced \ouralg, a distributed latent representation and neural swept-volume collision detector that predicts collision occurrence. By decomposing each object into a collection of locally optimized latent representations and pairing them with a neural decoder, our approach (i) captures only those surface features that matter for collision, (ii) exploits massive GPU parallelism through purely tensor operations, and (iii) generalizes across previously unseen shapes and trajectories by leveraging shape and temporal locality.

Experiments on swept-volume collision detection demonstrate that \ouralg\ deliver up to a \textbf{24× speed-up} over sphere-based GPU methods and more than a \textbf{100× speed-up} over mesh-based GJK—while simultaneously improving accuracy by \(\approx\) 5–15 \% even with the shapes unseen during training. When embedded in a trajectory-optimization–based motion planner, these gains result in higher success rates, lower penetration depths, and shorter planning times across tasks that span tight-tolerance assembly, dual-arm manipulation, and mobile manipulation in a cluttered mining tunnel. 

%% file: limitations.tex
\section{Limitation}


\subsection{Limitations in collision distance prediction}

Trajectory optimization methods, such as TrajOpt~\cite{schulman2013finding} and cuRobo~\cite{sundaralingam2023curobo}, rely on collision distance metrics that indicate penetration depth (negative values) during collisions and separation distances (positive values) when no collision occurs. In contrast, our method's function \(f_{SVCD}\) is trained primarily as a binary classifier, outputting collision probabilities. Therefore, we use logits—computed as the inverse sigmoid of these probabilities—as approximate surrogates for collision distances. While this approximation enables a smooth transition between collision and non-collision states and facilitates effective gradient computations during optimization, deviations from true collision distances can occasionally lead gradient-based trajectory optimizers to local minima.
In practice, a two-stage trajectory optimization approach combining MPPI and L-BFGS has proven effective at mitigating the issue of local minima, as validated through experiments in three challenging motion planning scenarios. This observation aligns with findings reported by cuRobo~\cite{sundaralingam2023curobo}.

\subsection{Application to raw sensory observations}
The current framework assumes access to the full mesh of the scene, including robot links, grasped objects and static environments. In practical real-world applications, obtaining such detailed representations is challenging. Recent advances in learning latent shape representations from raw sensory inputs, such as point clouds~\cite{irshad2022shapo} or RGB images~\cite{hong2023lrm}, indicate that it is possible to estimate object representation $\ourrepsim$ from these sources. This finding opens an exciting avenue for future research.


%% file: Appendix.tex
\clearpage




\section{Implementation Details in training}
\label{app:train_detail}

From given training dataset \(\mathcal{D}\), the training objective is \(\sum_{d\in\mathcal{D}}L_{SVCD}(d)\), which is the binary cross-entropy loss between predicted collision outcomes and ground-truth labels \(y_{SVCD}\). However, binary classification loss alone results in almost zero gradients when two objects are deeply penetrated into each other because the probability is close to 1 and nearly unchanged. This is undesirable because 
The training objective is defined as
\(
\sum_{d\in\mathcal{D}} \big[L_{SVCD}(d)\big],
\)
where \(L_{SVCD}\) is the binary cross-entropy loss between predicted collision outcomes and ground-truth labels \(y_{SVCD}\).
, and \(L_{reg}\) is a regularization loss given by:
\(
L_{reg}(d) = \Bigl(\lVert\nabla f_{SVCD}(d)\rVert - 1\Bigr)^2,
\)
encouraging smooth gradient variations beneficial for stable trajectory optimization \cite{gropp2020implicit}.

\section{Implementation Details in SVCD Accuracy}
\label{app:svcd_detail}
\textbf{Problem Generation.} First, we compute the shortest distance vector \(\delta\) between the moving robot and the fixed object by checking the penetration depth between the robot in a large number of intermediate configurations and the fixed object. The obstacle is then translated by \(\delta + \epsilon\), where \(\epsilon \sim \mathcal{N}(0, 0.03^{2})\) represents a small perturbation. After translating the fixed object, we calculate the penetration depth as we did previously to check the collision label. During problem generation, rejection sampling is used to balance positive and negative collision labels.

\textbf{Sphere Approximation.} To approximate the shape of a moving object with a set of spheres, we follow strategies from cuRobo. A set of spheres is constructed by combining spheres placed inside the object from voxelized interior regions with spheres centered at uniformly sampled surface points. For this process, we can vary the size of the voxel and the number of sampled surface points as hyperparameters.

\begin{table}[h]
\centering
\begin{tabular}{|c|c|c|}
\hline
\textbf{SVCD Method} & \textbf{Hyperparameter} & \textbf{Sampling Range} \\
\hline
\multirow{2}{*}{\shortstack{Convex Cell \\/ GJK}} 
& Trajectory Discretization Number & \( U(2, 256) \subset \mathbb{Z} \) \\
& Activation Length & \(U(-0.01, 0.01) \subset \mathbb{R} \) \\

\hline
\multirow{5}{*}{\shortstack{Sphere \\/ Sphere-Mesh \\ Distance}} 
& Trajectory Discretization Number & \( U(2, 64) \subset \mathbb{R} \) \\
& Activation Length & \(U(-0.01, 0.01) \subset \mathbb{R} \) \\
& Voxel Size (Sphere Approximation) & \(U(0.01, 1) \subset \mathbb{R} \) \\
& Number of Surface Points (Sphere Approximation) & \(U(0, 100) \subset \mathbb{Z} \) \\
& Mesh Simplification Voxel Factor & \(U(2, 32) \subset \mathbb{Z} \) \\
\hline
\multirow{2}{*}{\ouralg} 
& Trajectory Discretization Number & \( U(2, 100) \subset \mathbb{Z} \) \\
& Number of Collision Pairs After Broad-Phase Filtering & \(U(2, 256) \subset \mathbb{Z} \) \\

\hline
\end{tabular}
\caption{Hyperparameter Sampling Ranges for SVCD Baselines. $U$(min, max) denotes uniform distribution}
\label{tab:hyperparam_ranges}
\end{table}

\begin{algorithm}[htbp]
\caption{Proposed SVCD Algorithm}
\label{alg:refined_ccd_algorithm}
\begin{algorithmic}[1]
\Require Static Mesh: $\text{mesh}_{static}$, Moving Mesh: $\text{mesh}_{mov}$, Trajectory: $\tau$
\Ensure Collision prediction output: $\text{ccd\_output}$

\State $\ourrepsim_{static}, \ourrepsim_{mov} \gets \text{encode}(\text{mesh}_{static}, \text{mesh}_{mov})$
\State $\{(p_1^{(i)}, r_1^{(i)}, z_1^{(i)}, p_2^{(i)}, r_2^{(i)}, z_2^{(i)}), t^{\dagger(i)}\}_{i=1}^M \gets \text{broadPhase}(\ourrepsim_{static}, \ourrepsim_{mov}, \tau)$

\State $\text{SVCD\_output} \gets -\infty$
\For{$i = 1$ \textbf{to} $M$}
    \State $\xi^{(i)}, \tau(t^{\dagger(i)}) \gets \text{LinearApproximation}(\tau, t^{\dagger(i)})$
    \State $\text{SVCD\_logit}^{(i)} \gets f_{SVCD}(p_1^{(i)}, p_2^{(i)}, z_1^{(i)}, z_2^{(i)}, \xi^{(i)}, \tau(t^{\dagger(i)}))$
    \State $\text{SVCD\_output} \gets \max(\text{SVCD\_output}, \text{SVCD\_logit}^{(i)})$
\EndFor

\Return $\text{SVCD\_output}$
\end{algorithmic}
\end{algorithm}

\section{Domain Description for Motion Planning}
\label{app:mp_domain}

\textbf{Dish insertion:} A fixed‐base UR5 must insert three distinct dishes (sampled without replacement from six shapes) into a rack. We randomly place the rack in a collision‐free pose and sample a collision‐free robot start configuration. Hard-coded target poses specify both the end effector and the final location of each dish. After each insertion, physics is enabled to let the dish settle; once settled, it becomes a static obstacle for subsequent insertions.

\textbf{Bimanual insertion:} The ARMADA system, with two 6 DoF arms, performs a peg‐into‐slot assembly. We generate collision-free start and goal configurations via rejection sampling: for each trial, we sample a peg pose, grasp pose, and peg–hole shape, solve the inverse kinematics for joint angles, and discard any samples that collide.

\textbf{Mining site navigation:} A UR5 on a mobile base carries a pickaxe through a synthetic mining tunnel, avoiding wagons and beams. In each episode, we randomly select four meshes from eight obstacle candidates and place them at fixed scene locations. The robot’s start configuration is sampled from a bounded, collision‐free region, while the pickaxe’s goal location remains constant.


\section{Implementation Details in Motion Planning}
\label{app:mp_detail}

\begin{figure}
    \centering
    \includegraphics[width=1\linewidth]{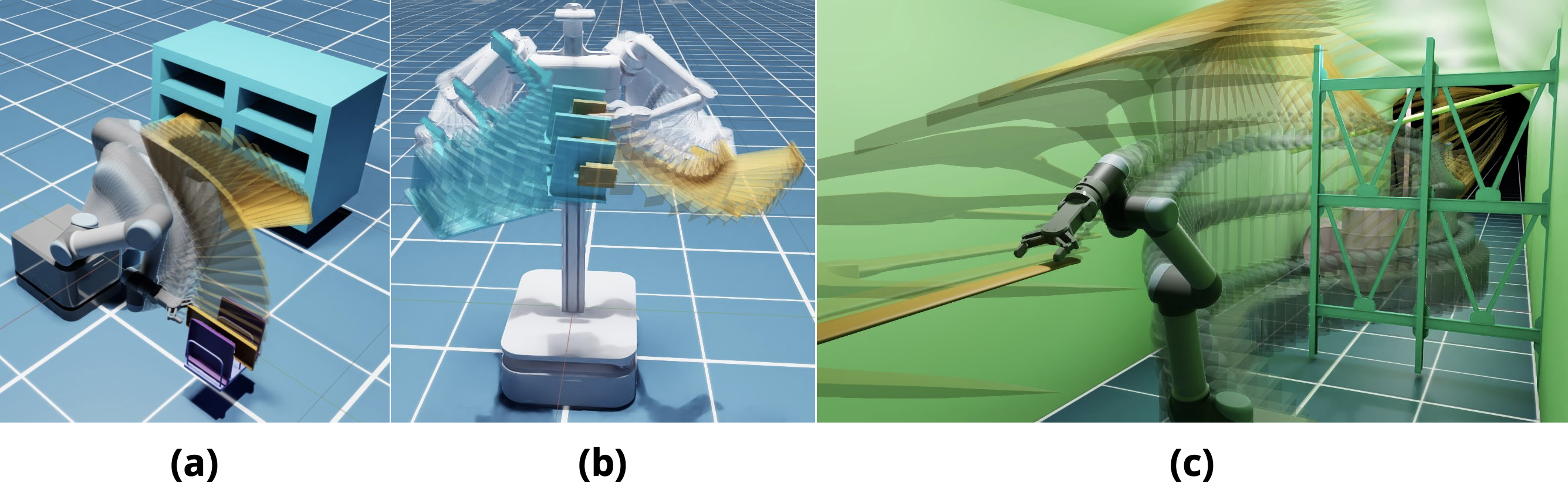}
    \caption{Illustration of three robotic tasks solved by our proposed algorithm. (a) Dish insertion: A UR5 robot precisely inserts dishes into a dish rack. (b) Peg assembly: ARMADA~\cite{kim2025design}, a 12-DOF bimanual manipulator, simultaneously holds a peg in one arm and a slot in the other, accurately assembling the peg into the designated slot. (c) Mobile manipulation at a mine tunnel: A mobile manipulator transports a pickaxe to their target location, carefully navigating around obstacles such as beams and a wagon. Each task highlights the critical importance of precise, collision-free motion planning.}
    \label{fig:mp_task_visualization}
\end{figure}

To demonstrate improvements in motion planning performance, we incorporate \ouralg\ into a trajectory optimization framework modeled after cuRobo~\cite{sundaralingam2023curobo}. In this framework, trajectories are parameterized using splines with a fixed number of control points. Our method adopts a two-stage optimization process. First, a particle-based solver (MPPI~\cite{williams2016aggressive}) is employed to promote exploration, and then the trajectory is refined using L-BFGS~\cite{liu1989limited}.

Initially, we represent the trajectory as a spline with a fixed number of control points $\tau_{control}\in\mathbb{R}^{M \times F}$, where $M$ is a number of control points, and $F$ is a number of actuation of the robot. We then apply an optimization-based motion planner with an objective function from cuRobo but with custom collision cost:
\vspace{-1mm}
\begin{equation}
    C_{traj}(\tau_{control})= C_{smooth}(\tau_{control}) + \alpha_{col}C_{SVCD}(\tau_{control})
    \label{eq:curobo_loss}
\end{equation}
\vspace{-1mm}
where $C_{SVCD}$ and $\alpha_{col}$ denotes SVCD cost computed using \ouralg\ and its coefficient. For $C_{smooth}=\alpha_{vel}C_{vel}+\alpha_{acc}C_{acc}+\alpha_{jerk}C_{jerk}$, where each term indicates velocity, acceleration, and jerk minimization cost with pre-defined coefficient.
For cuRobo-\ouralg\ and cuRobo-\ouralg-discrete, we modify the collision cost term in the loss function \eqref{eq:curobo_loss} by replacing it with collision logits computed by \(f_{ccd}\). For each task, we use different hyperparameters. See Table \ref{tab:bimanual-hyperparams}, \ref{tab:mining-hyperparams}, \ref{tab:dish-insertion-hyperparams}. We also modify the calculating gradient during L-BFGS to be the bundled gradient~\cite {suh2022bundled}, which combines gradient evaluations from multiple samples or time steps into a single aggregated update to improve optimization efficiency and stability, benefiting all baselines.

\section{Applying transformation to the latent vectors}

Another critical consideration in designing latent representations and the neural SVCD decoder is the application of rigid-body transformations. During collision evaluation, each robot link must be positioned according to specific trajectory configurations, requiring transformations described by SE(3)—a combination of rotations and translations. Using explicit mesh representations, applying transformations is straightforward: one simply applies transformations directly to mesh vertices before checking for collisions. However, when using latent representations $z$ for robot links, applying SE(3) transformations is not trivial, as latent representations do not inherently support direct geometric transformations.

One option is to use neural network-based transformation operators~\cite{qi2017pointnet}, which receive both the transformation and $z$ and output the transformed $z$. However, they do not guarantee the rigid-body transformation properties required. For instance, applying a 180° rotation about the z-axis twice should reproduce the original representation, a consistency that such operators cannot reliably ensure.

To overcome these limitations, we introduce an analytical transformation operation within a high-dimensional latent space that explicitly preserves rigid transformation properties.
Applying a rigid-body transformation \( T \in SE(3) \) analytically involves separately treating translation and rotation. Given \( T = (R, t) \), with \( R \in SO(3) \) being a rotation matrix and \( t \in \mathbb{R}^3 \) a translation vector, we define the transformation of each point-latent pair as:
\(
T \cdot (p_i, z_i) := (R \cdot p_i + t,\; D(R) \cdot z_i)
\)
where \( D(R) \in \mathbb{R}^{K \times K} \) is a rotation operator applied directly within the latent space, explicitly constructed following the equivariant rotation formulation proposed in~\cite{son2024an}. This operator \( D(R) \) ensures that latent representations transform consistently with geometric rotations, preserving rigid-body properties. Intuitively, this means translations shift only the spatial coordinates \( p_i \), while rotations alter the latent encodings \( z_i \), effectively rotating the local geometric feature descriptions encoded within each latent vector.

\section{Implementation of encoder and neural SVCD decoder}

\begin{figure}
    \centering
    \includegraphics[width=0.5\textwidth]{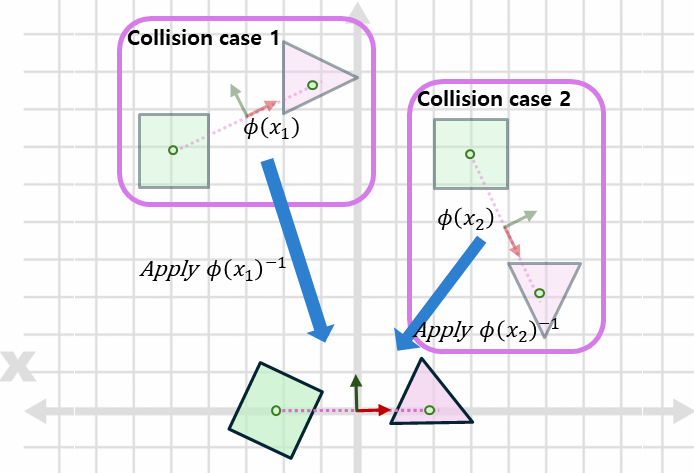}
    \caption{Illustration of pre-processing for achieving invariance. Case 1 and 2 have the same relative transform between two objects, but their global transforms are different. If we treat the two objects as a single composite rigid body, we can assign frames $\{\phi(x_1)\}$ and $\{\phi(x_2)\}$ whose origin is at the mid-point of the centers of two objects, and whose direction is determined by the line intersecting the centers. We then apply $\phi(x_1)^{-1}$ and $\phi(x_2)^{-1}$ to these frames so that they are at the origin of the world frame, with their orientation aligned with that of world frame as shown in the bottom. This preprocessing step ensures consistent input irrespective of the objects' global poses.}
    \label{fig:preprocessing_decoder}
\end{figure}
For any two meshes and their corresponding moving direction \(\xi\), the SVCD result is invariant under both SE(3) transformations and scaling. In other words, if a consistent transformation \(T\) is applied to the meshes and the moving direction or if a uniform scale \(s\) is applied, the collision detection outcome remains unchanged. We leverage this property when designing our neural CCD decoder.
The inputs to the neural CCD decoder, \(f_{SVCD}\), include the latent representations \(z_i\), positions \(p_i\), transformations \(T_i\), and twists \(\xi_i\) (which contain both linear and angular velocities) at critical collision times determined during the broad-phase.

Building on ideas from \cite{son2024an, puny2022frame, yang2024equivact}, we first preprocess these inputs to achieve invariance with respect to additional transformations and scaling. The preprocessing pipeline is illustrated in Figure \ref{fig:preprocessing_decoder}.
Let the combined input be \(x = [\xi; z_1; z_2; p_1; p_2]\).
We define the preprocessing function as 
\[\Phi(x)=\frac{1}{max(\lVert z_1\rVert,\lVert z_2\rVert)}(\phi(x)^{-1}\cdot x),\]
with $\phi: \mathbb{R}^{M} \to SE(3)$ such that $\phi(T \cdot x)=T\cdot\phi(x) \quad \forall T\in SE(3)$, where \(M\) is the dimension of \(x\) and $\lVert z \rVert$ is Euclidean norm of $z$. It is straightforward to verify that \(\Phi\) is invariant under any transformation in \(SE(3)\) and scale.
We define \(\phi\) analytically by setting its translation component to \(-(p_1+p_2)\) and aligning its rotation \(R\in SO(3)\) with the line connecting \(p_1\) and \(p_2\), thereby rigidly attaching the transformation to the pair of bodies. This construction ensures the property \(\phi(T \cdot x)=T\cdot\phi(x)\) holds.

The preprocessed inputs are then passed through multiple multilayer perceptrons (MLPs). The network outputs a scalar logit that represents a binary collision prediction (with positive logits indicating a collision). Finally, after predicting collisions for every \(p_i\) and \(p_j\) pair identified during the broad-phase, we apply max pooling to obtain the final logits indicating a collision between the rigid bodies.

The shape encoder processes an input point cloud to produce $N$ local representations $\{(z_i, p_i)\}_{i=1}^N$. The sample positions $\{p_i\}$ are selected via Furthest Point Sampling (FPS), and the corresponding latent vectors $\{z_i\}$ are learned by a neural network. Concretely, every surface point is assigned to its nearest sample $p_i$, and all points in each partition are fed—independently but using shared weights—through the encoder to generate the local $z_i$. To ensure compatibility with our latent-space transformations, we adopt the FER-VN-OccNet encoder architecture from \cite{son2024an}.

\begin{table}[ht]
\centering
\begin{tabular}{lcccc}
\toprule
\textbf{Hyperparameter} &
  \shortstack{cuRobo-\\sphere-\\100} &
  \shortstack{cuRobo-\\sphere-\\1200} &
  \shortstack{cuRobo-\\\ouralg} &
  \shortstack{cuRobo-\\\ouralg-\\discrete} \\
\midrule
Number of initial seeds & \multicolumn{4}{c}{2}                \\
MPPI - interpolation number & \multicolumn{3}{c}{4}    & 100  \\
MPPI - iteration & \multicolumn{4}{c}{40}               \\
MPPI - number of samples & \multicolumn{3}{c}{200}  & 50   \\
MPPI - number of control points & \multicolumn{4}{c}{5}            \\

LBFGS - interpolation number & \multicolumn{3}{c}{2}    & 100  \\
LBFGS - iteration & \multicolumn{4}{c}{10}               \\
LBFGS - number of control points & \multicolumn{4}{c}{10}           \\

$\alpha_{vel}$                       & \multicolumn{4}{c}{10.0}              \\
$\alpha_{col}$                       & \multicolumn{4}{c}{1.0}               \\
$\alpha_{acc}$               & \multicolumn{4}{c}{2}                 \\
$\alpha_{jerk}$              & \multicolumn{4}{c}{10}                \\


Activation distance   & \multicolumn{2}{c}{0.010} & - & -    \\
Number of sphere for robot & \multicolumn{2}{c}{891} & - & - \\
Number of sphere for grasping object & 100 & 1200 & - & - \\
\bottomrule
\end{tabular}
\caption{Hyperparameter settings for the bimanual insertion task across motion planners.}
\label{tab:bimanual-hyperparams}
\end{table}

\begin{table}[ht]
\centering
\begin{tabular}{lcccc}
\toprule
\textbf{Hyperparameter} &
  \shortstack{cuRobo-\\sphere-\\50} &
  \shortstack{cuRobo-\\sphere-\\2000} &
  \shortstack{cuRobo-\\\ouralg} &
  \shortstack{cuRobo-\\\ouralg-\\discrete} \\
\midrule
Number of initial seeds & \multicolumn{4}{c}{2}                \\
MPPI - interpolation number & \multicolumn{3}{c}{3} & 100 \\
MPPI - iteration               & \multicolumn{4}{c}{40}               \\
MPPI - number of samples       & \multicolumn{3}{c}{400} & 50              \\
MPPI - number of control points     & \multicolumn{4}{c}{8}   \\

LBFGS - interpolation number & \multicolumn{3}{c}{8} & 100   \\
LBFGS - iteration              & \multicolumn{4}{c}{10}               \\
LBFGS - number of control points    & \multicolumn{4}{c}{3}                \\

$\alpha_{vel}$  & \multicolumn{4}{c}{1}               \\
$\alpha_{col}$  & \multicolumn{4}{c}{5}               \\
$\alpha_{acc}$  & \multicolumn{4}{c}{10}               \\
$\alpha_{jerk}$ & \multicolumn{4}{c}{50}               \\


Activation distance   & 0.040 & 0.040 & -   & -   \\
Number of sphere for robot & \multicolumn{2}{c}{397} & - & - \\
Number of sphere for grasping object      & 50  & 2000 & -   & -   \\
\bottomrule
\end{tabular}
\caption{Hyperparameter settings for the mining site navigation task across motion planners.}
\label{tab:mining-hyperparams}
\end{table}

\begin{table}[ht]
\centering
\begin{tabular}{lcccc}
\toprule
\textbf{Hyperparameter} &
  \shortstack{curobo-\\sphere-\\50} &
  \shortstack{curobo-\\sphere-\\400} &
  \shortstack{curobo-\\\ouralg} &
  \shortstack{curobo-\\\ouralg-\\discrete} \\
\midrule
Number of initial seeds & \multicolumn{4}{c}{2}                      \\
MPPI - interpolation number    & \multicolumn{3}{c}{4}   & 50              \\
MPPI - iteration              & \multicolumn{4}{c}{40}                     \\
MPPI - number of samples              & \multicolumn{3}{c}{200} & 50              \\
MPPI - number of control points & \multicolumn{4}{c}{5}                      \\

LBFGS - interpolation number  & \multicolumn{3}{c}{2}   & 50              \\
LBFGS - iteration               & \multicolumn{4}{c}{10}                     \\
LBFGS - number of control points & \multicolumn{4}{c}{10}                     \\

$\alpha_{vel}$ & \multicolumn{4}{c}{10.0}                  \\
$\alpha_{col}$ & \multicolumn{4}{c}{1.0}                   \\
$\alpha_{acc}$ & \multicolumn{4}{c}{2}                     \\
$\alpha_{jerk}$ & \multicolumn{4}{c}{10}                    \\


Activation distance    & 0.040 & 0.040 & -                   & -               \\
Number of sphere for robot & \multicolumn{2}{c}{397} & - & - \\
Number of spheres for grasping object & 50    & 400    & -                   & -  \\
\bottomrule
\end{tabular}
\caption{Hyperparameter settings for the dish insertion task across motion planners.}
\label{tab:dish-insertion-hyperparams}
\end{table}